\documentclass{article}
\usepackage{spconf,amsmath,graphicx}
\usepackage{graphicx}
\usepackage{amsmath}
\usepackage{amssymb}
\usepackage[algo2e]{algorithm2e}
\usepackage[tight,footnotesize]{subfigure}
\usepackage{multicol}

\usepackage{url}
\usepackage{array}
\usepackage{epsfig}
\usepackage{algorithm}
\usepackage{subfigure}
\usepackage{multicol}
\usepackage{color}

\def \la {\leftarrow}

\def \L {{\bf L}}
\def \S {{\bf S}}

\def \0 { {\bf 0} }

\def \X {{\bf X}}

\def \f  {{\bf X}^{(t)}}

\def \R{\mathcal R}
\def  \A {\mathcal A}

\def \y {{\bf Y}^{(t)}}

\def \T {\mathcal T}
\def \Tr {\bf \mathbb{S}}
\title{Low-Rank and Sparse Matrix Decomposition with a-priori knowledge  for Dynamic 3D MRI reconstruction } %to minimize an  $\ell_p$-norm  }
\name{Author(s) Name(s)}
\name{Dornoosh Zonoobi \qquad Shahrooz Faghih Roohi  \qquad Ashraf A. Kassim $^{\star}$}

\address{Department of Electrical \& Computer Engineering \\National University of Singapore\\
	Singapore}

\begin{document}
\topmargin=0mm
%\ninept
%
\maketitle
\begin{abstract}
It has been recently shown that incorporating priori knowledge significantly improves the
performance of basic compressive sensing based approaches. We have managed to successfully
exploit this idea for recovering a matrix as a summation of a Low-rank and a Sparse component from compressive measurements. 
When applied to the problem of construction of 4D Cardiac MR image sequences in real-time from highly under-sampled $k-$space data, our proposed method achieves superior reconstruction quality compared to the other state-of-the-art methods.

%n this paper, we propose a Compressive Sensing based approach to the problem of r Our proposed method is able to extract useful priori information and incorporate it into a modified iterative soft-thresholding-based  algorithm for fast  reconstruction of 3D dynamic MR volumes from highly undersampled k-space data. 

 \end{abstract}

%%%%%%%%%%%%%%%%%%%%%%%%%%%%%%%%%%%
%%%%%%%%%%%%%

%%%%%change

\section{\uppercase{Introduction}}
\label{sec:introduction}

\begin{figure*}[htp]
\centering
{\includegraphics[trim=19mm  0mm 20mm 0mm, clip,width=0.6\textwidth]{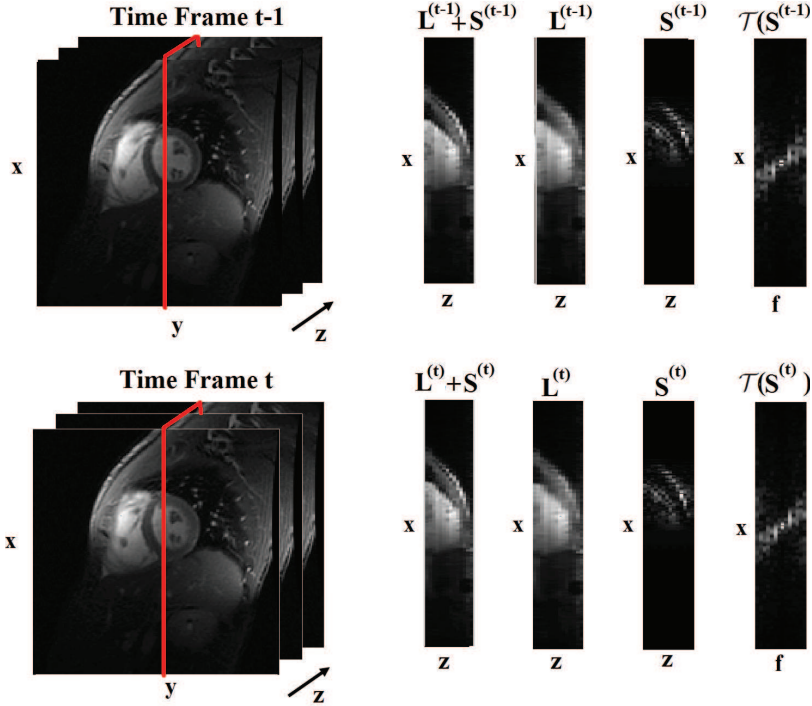}}
\caption{Illustration of L+S decomposition of fully-sampled 3D cardiac cine at time $t$ and $t+1$. }
\label{fig:LS}
\end{figure*}

A fundamental problem in dynamic MRI, such as real-time cardiac MRI (rtCMR), is the limitation of spatial and temporal resolution which is due to the slow data acquisition process of this modality. 
This problem is even more profound when dealing with 4D MR volumes. Compressive Sensing (CS) has been shown to be able to overcome these challenges and recover  MRI images  from much smaller $k$-space measurements than conventional reconstruction methods. To achieve this, earlier CS-based methods assumed that the MRI images have a sparse  representation in some  known transform domain \cite{zonoobi2014dependent,Hu2012,Camera1,dgini} and the idea was easily extended to the reconstruction of dynamic MRI  images data  by jointly reconstructing the entire sequence by treating it as higher dimensional data  \cite{bookmri,our}. In other works, the high spatiotemporal correlation was utilized  to recover dynamic images  by solving a low rank matrix completion problem in which 
 each temporal frame is a column of the recovered matrix  \cite{Zhao2010}, \cite{Haldar2011}. 
 
 Some studies have reported much improved results that were obtained by combining rank deficiency and transform domain sparsity. These include proposals to recover the image as a solution which is both sparse and low rank \cite{Majumdar2012}, \cite{Gao2012}; and other proposals  that decompose the data in two low-rank ($\L$) and sparse ($\S$) components \cite{Majumdar2012_2}, \cite{Gao2012,Goud2010}, where $\L$  models the correlated information between frames and $\S$ represents the rapid change of data over time. 
 
 More recently, it has been shown that incorporation of the priori knowledge into the reconstruction  of sparse signals can significantly improve their performance \cite{Zonoobi2013,modcs,zonoobi2014ecg}.   This idea have been used in the  Modified-CS  \cite{Vaswani2010} to  {\it recursively} reconstruct a time sequence of MRI images in real-time from highly under sampled measurements by using a-priori knowledge obtained from the previous reconstructed image. The Modified-CS  uses the support of the previous time instance as a partially known part of the current  support and finds a signal which satisfies the observations and is sparsest outside the support of the previous time instant. The a-priori based methods,  only model the signal of interest as one sparse component.  However, it is observed  that the $\L$ and $\S$ decomposition can model dynamic MRI data significantly better than a low-rank or a sparse model alone, or than a model in which both constraints are enforced simultaneously \cite{Otazo2013,APSIPA,Shahab}.
To the best of our knowledge, no previous work has been done to incorporate the priori information into image recovery while the image is modelled as a summation of  a low rank and a sparse component. 

In this paper we first propose a re-formulation of the $\L$ and $\S$ decomposition to take into account  some priori knowledge and then use  a  soft-thresholding based algorithm to efficiently solve it. The algorithm is then employed to reconstruct a time sequence of 3D  cardiac MRI volumes  from highly undersampled measurements.

The rest of this paper is organized as follows:  this section ends with a description of the notations used. Section 2 presents the problem of reconstruction of 3D dynamic MRI volumes  and the current state-of-the-art CS-based approaches that
address this problem. In section 3, we provide details of our proposed algorithm which we call {\it Priori L+S}. Finally, we present and analyze our experimental results in section 4 before providing the concluding remarks in section 5.

{\bf Notations:} Throughout the paper, matrices are denoted by  boldface letters (e.g.  $\X, \S $) while scalars are shown by small regular letters (e.g. $ n, m, k, r$) and linear maps and operators are denoted by bold calligraphic  uppercase letters ($\T, \A, {\Sigma}$) and  $\A^{-1}$ denotes the adjoint of the operator. Superscript $(t)$ added to a matrix refers to that of time $t$. For a matrix, the notation $\bf {M} |_{\mathcal S}$ forms a sub-matrix that contains elements with indices in ${\mathcal S}$.

\section{Problem Formulation}
\label{PA}
 
The low rank and sparse matrix decomposition (L+S) is particularly suitable to the problem of dynamic imaging, where the low rank component models the temporally correlated background and the sparse component represents  the dynamic information that lies on top of the background \cite{Gao2012,Lingala2011}.   To apply the low-rank and sparse matrix  decomposition to 3D dynamic MRI, lets assume that the 3D volume of interest is of size $[n_x\times n_y\times n_z], (n_x,n_y>n_z) $ which is changing with time. 
At each time instance $t$ the 3D volume  is converted to a matrix $\f \in \R^{(n_x n_y) \times n_z} $,  where each column is consist of a frame.  This matrix could be then decomposed into a low rank  matrix $\L^{(t)}$ and a matrix $\S^{(t)}$, which we assume to have a sparse representation in some known basis $\T$ (such as Wavelets \cite{kassim2008}), as  
$
\f= \L^{(t)}+\S^{(t)}
$.

Figure \ref{fig:LS} shows a cross section of the low rank and sparse components  of cardiac data sets for two adjacent time instances ($t$ and $t+1$). It can be seen that 
$\L^{(t)}$  represents the background component and $\S^{(t)}$ corresponds to the changes from a frame to another, e.g., organ motions or contrast -enhancement, etc \cite{Majumdar2012,Zonoobi2014,Feng2012}. 

With this the problem can be posed as follows: let $\A$ be the acquisition/sampling operator that performs a frame-by-frame k-space under-sampling  of the $t^{th}$ volume ($\A: \R^{(n_x n_y) \times n_z}\rightarrow \R^{m\times n_z}$, where $m\ll n_xn_y$). Using this operator, the under-sampled acquisition of $\f$ can be expressed  as: $$ {\bf d}^{(t)}= \A( \f)+\eta $$ where $\y$ is the observation matrix of size $m\times nz$, and is assumed
to be incoherent with respect to the sparsity basis. Also $\eta$ is the
measurement noise with finite energy (i.e. $\|\eta\|_2\le \epsilon_1$), which can be modelled
as a complex Gaussian noise.  The problem, at each time instance $t$, is then to recover the original $ {\f}$, from the corresponding compressive samples $\y$ assuming that the signal of interest is can be decomposed into  low rank and sparse components. 
The problem of recovering each volume from the compressive measurements   can be then formulated as:
\begin{align}
\label{Eq_PCA}
 ( \L^{(t)}, \S^{(t)})= & {\arg\min } \{\|\Sigma({{\bf L^{(t)}}})\|_0+   \|{{\T(  \S^{(t)})}}\|_0 \} \\  
 & {\rm subject\; \;  to} \;  \;  \|  \y- \A (  {\L^{(t)}}+  {\S^{(t)}} ) \|_{2} \le \epsilon_1,  \nonumber
\end{align}
where  $\T$ is a sparsifying transform for $\S$, and $\Sigma$ is an operator that maps any matrix to the vector of its singular values (i.e. $\|\Sigma({{\bf L^{(t)}}})\|_0 =rank( {{\bf L^{(t)}}})).$

Solving the above minimization problem is known to be computationally unwieldy in view of its combinatorial nature. As a consequence, we are compelled to resort to an alternative convex approximation as follows:
\begin{align}
\label{Eq_PCA2}
 ( \L^{(t)}, \S^{(t)})= & {\arg\min } \{ \|\Sigma({{\bf L^{(t)}}})\|_1 + \|{{\T( \bf S^{(t)})}}\|_1\}  \\  
  & {\rm subject\; \;  to} \;  \;  \|  \y- \A (  {\L^{(t)}}+  {\S^{(t)}} ) \|_{\ell_2} \le \epsilon_1, \nonumber
\end{align}
where  $\|\Sigma({{\bf L^{(t)}}})\|_1$ is the nuclear norm of ${\bf L^{(t)}}$.
This convex problem can be solved efficiently using an iterative algorithm, thereafter referred to as  L+S method \cite{Otazo2013}, which is closely related to \cite{Beck2009} and  \cite{Cai2010} for sparse ($\T(  {\S})$) and low-rank matrix recovery (${\L}$), respectively. The L+S method  starts from a signal proxy and then at each iteration proceeds through  three steps to update its estimates of the low rank matrix  and the sparse component using a soft-thresholding operator. This operator is defined as:
$${\bf \mathbb{S}}\{x,\lambda\}= \frac{x}{|x|}max\left (|x|-\lambda,0 \right) $$ in which x could be a complex number and the threshold $\lambda$ is real valued. This is extended to matrices by applying it to each element of that matrix. 

It is known from the  literature that recovery of sparse vectors and low-rank matrices can be accomplished when the measurement operator $\A$ satisfies the appropriate RIP or RRIP conditions \cite{PCA1}. The above formulation, however, does not take into account any priori information that may be available about the low rank/sparse components. 

\section{Low-Rank and Sparse Matrix Decomposition with a-priori information}
\label{sec:2}
 \begin{figure*}
\centering
{\includegraphics[trim=10mm  10mm 10mm 10mm, clip=true,width=0.9\textwidth]{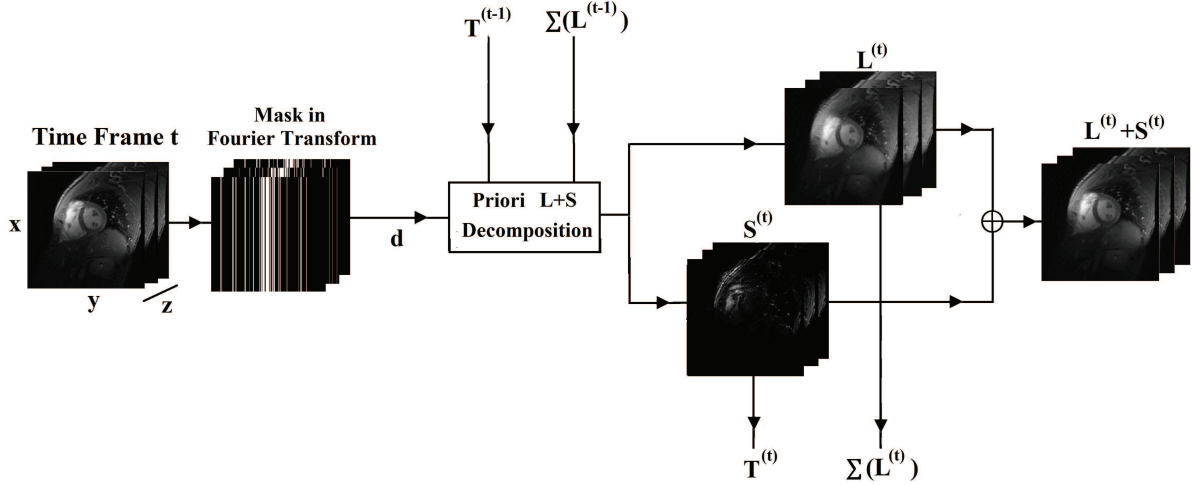}}
\caption{ Overview of the priori L+S scheme. }
\label{fig:block}
\end{figure*}

To  reconstruct images from even  fewer number of samples than L+S method \cite{Otazo2013}, we aim to use  the $L$ and $S$ components of the previously reconstructed volume to guide the reconstruction of the current time volume.  The idea is based on the observation that the $\L$ and $\S$ components of  each MRI volume is very closely related to those of the adjacent time instances.  This is not surprising as it is known that  dynamic images are highly redundant in space and time \cite{8}.  To illustrate  this, figure \ref{fig:LS}  shows a cross-section of the low rank and sparse components of a fully-sampled cardiac data set for two adjacent time instances. From the figure it can be seen that $\L^{(t-1)}$ and $\L^{(t)}$ are quite similar, in fact $\|\Sigma (L^{(t)})-\Sigma(L^{(t-1)})\|_2 =0.04$. This means that  vector of singular values of  $\L^{(t-1)}$ and $\L^{(t)}$ are very close in Euclidean space.  Similarly, support of $\T(\S^{(t)})$  is much the same as the one of $\T(\S^{(t-1)})$. In this case, for instance,  the support change turns out to be less than 5\% of the support size. Therefore, support of $\T(\S^{(t-1)})$ can be viewed as an  a-priori knowledge of the partial support  of $\T(\S^{(t)})$. 
 Based on the above observations, to recover $\f$, we modify the formulation of the problem to incorporate the information of $\T(\S^{(t-1)})$ and $\L^{(t-1)}$  as follows: 
  \begin{align}
\label{Eq_PCA3}
 ( \L^{(t)}, \S^{(t)})= & {\arg\min } \{ \| \Sigma({{\bf L^{(t)}}})\|_1 + \|{{\T( \bf S^{k})}} |_{\bar T^{(t-1)}}\|_1\}  \nonumber \\  
  & {\rm subject\; \;  to} \;  \;  \|  \y- \A (  {\L^{k}}+  {\S^{k}} ) \|_{2} \le \epsilon_1, \nonumber\\& \| \Sigma({{\bf L^{k}}})- \Sigma({{\bf L^{k-1}}})\|_2 \le \epsilon_2 
\end{align}
 where $T^{(t-1)}$ denotes the support of $\T(\S^{(t-1)})$ and $\bar T^{(t-1)}$ is the complement of $T^{(t-1)}$.
Basically we are searching for an image which satisfies the observations, its $\S$ component is sparsest outside $T^{(t-1)}$ and at the same time it has  $\Sigma(\L^{(t)})$   closest to $\Sigma(\L^{(t-1)})$.

The above formulation is convex and therefore it has a unique solution, however using convex-based optimization methods may not be practical for large-scale problems  due to their considerable computational complexity and memory requirements \cite{Cosamp}. 
Therefore we solve (\ref{Eq_PCA})  using an iterative algorithm inline with  L+S method  \cite{Otazo2013}.
\begin{algorithm}[ht]
\label{alg:gr1}
\KwIn {$\y , \A,\S^{(t-1)},\L^{(t-1)},\lambda_T,\lambda_S $} 
(0) Initialization:  $\X_0=\A^{-1}(\y),  \S_0=0;$ \\ {
\While{ not converged  }{	
\vspace{0.05in}
(1) {\small Singular-value soft-thresholding of $\L$: }\\
${\bf L}_{it-1}={\bf X}_{it-1}-{\bf S}_{it-1}$;\\
$   \L_{it}= \Sigma^{-1}(\Tr\{$
$\Sigma({\bf L}_{it-1}),\lambda_L\});$\\\\
(2) {\small   Imposing the priori knowledge on $\L$: }\\
$D_{it}=(\Sigma (\L_{it})-\Sigma(\L^{(t-1)}))$;\\
$   \L_{it}=\Sigma^{-1}(\Sigma (\L_{it})-\lambda_p D_{it}) $\\\\
(3) {\small  Imposing Sparsity and the priori knowledge on S: }\\
$T^{(t-1)}=supp(\T(\S^{(t-1)}))$;\\
$   \S_{it}= \T^{-1} ( \Tr\{\T(\S_{it-1})|_{\bar T^{(t-1)}},\lambda_S\}$; \\ \\
(4) {\small   Update estimation to minimize error:} \\
$E_{it}=  \y- \A (  \L_{it}+  \S_{it})$;\\
$\X_{it}=  \L_{it}+  \S_{it}-\A^{-1}(E_{it})$;}
\KwOut{$  \L^{(t)} \la   \L_{it},   \S^{(t)}  \la   \S_{it}$}}
\caption {PrioriL+S decomposition. }
\end{algorithm}

Our proposed algorithm, which is summarized below, mainly differs with \cite{Otazo2013} in two steps where we impose the available priori-knowledge into estimation of $\S$ and $\L$ components. 
{\bf Initialization:} similar to the original L+S algorithm, we start with an initial estimation of $\f$ and we set the sparse component to be all zeros.

{\bf Singular-value soft-thresholding:}  to impose the low-rank property on $\L^{(t)}$, in this step at the $i-th$ iteration the vector of singular values of ($\X_{it-1}-\S_{it-1}$) is soft thresholded.

{\bf Imposing the priori knowledge on $\L$:} this step is designed to imposed the available priori knowledge of the low rank component $\L^{(t)}$  which is extracted from $\L^{(t-1)}$. To this end at each iteration $it$, it minimizes the  Euclidean distance between the singular values of $\L_{(it)}$ and the previously reconstructed component, $\L^{(t-1)}$, by moving into its gradient decent direction. 

{\bf  Imposing Sparsity and the priori knowledge on $\S$ :}  In this step the goal is only force $\T(\S_{it})$  to be sparse in locations not belonging to the spikes of the previous time instance ($T^{(t-1)}$). To this end, the algorithm only shrinks  those elements not belonging to $T^{(t-1)}$. 

{\bf  Update estimation to minimize error:}  the new $\X$ is finally obtained by enforcing measurement  consistency, where the aliasing artifacts corresponding to the residual in k-space  are subtracted from $L_{it}+S_{it}$. The algorithm iterates until the relative change in the solution is less than $10^{-3}$. 

Figure \ref{fig:block} shows the priori $L+S$ scheme for reconstructing the entire time sequence. At each time $t$, Algorithm 1 is used to recover $\f$ except for $t=1$, where the
simple $L+S$ algorithm is used as no priori knowledge is available for the reconstruction of the first volume.

\begin{figure}[!h]
\centering
{\includegraphics[trim=20mm 20mm 10mm 1mm, width=0.2\textwidth]{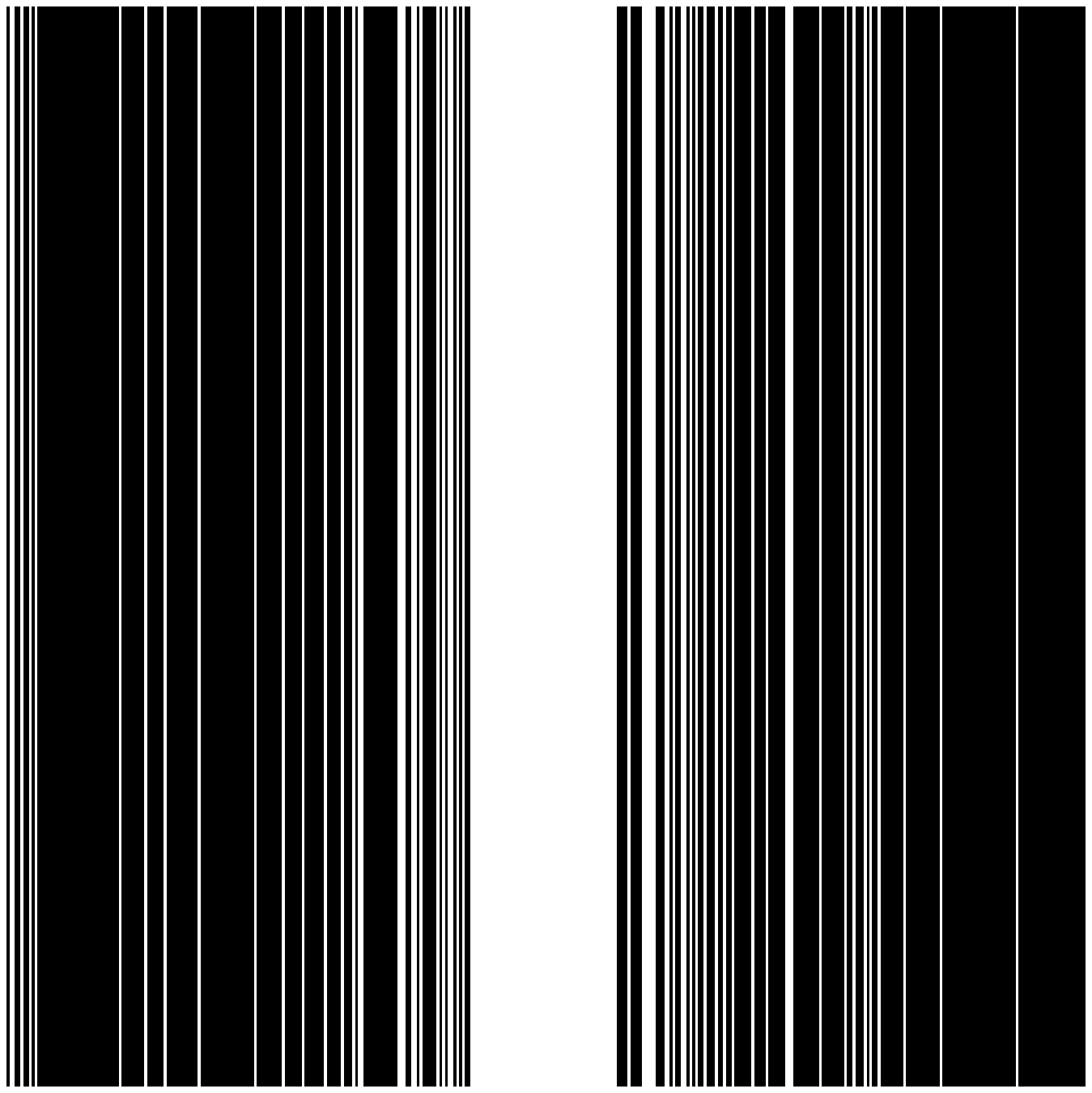}}
{\includegraphics[trim=20mm  20mm 10mm 0mm, width=0.2\textwidth]{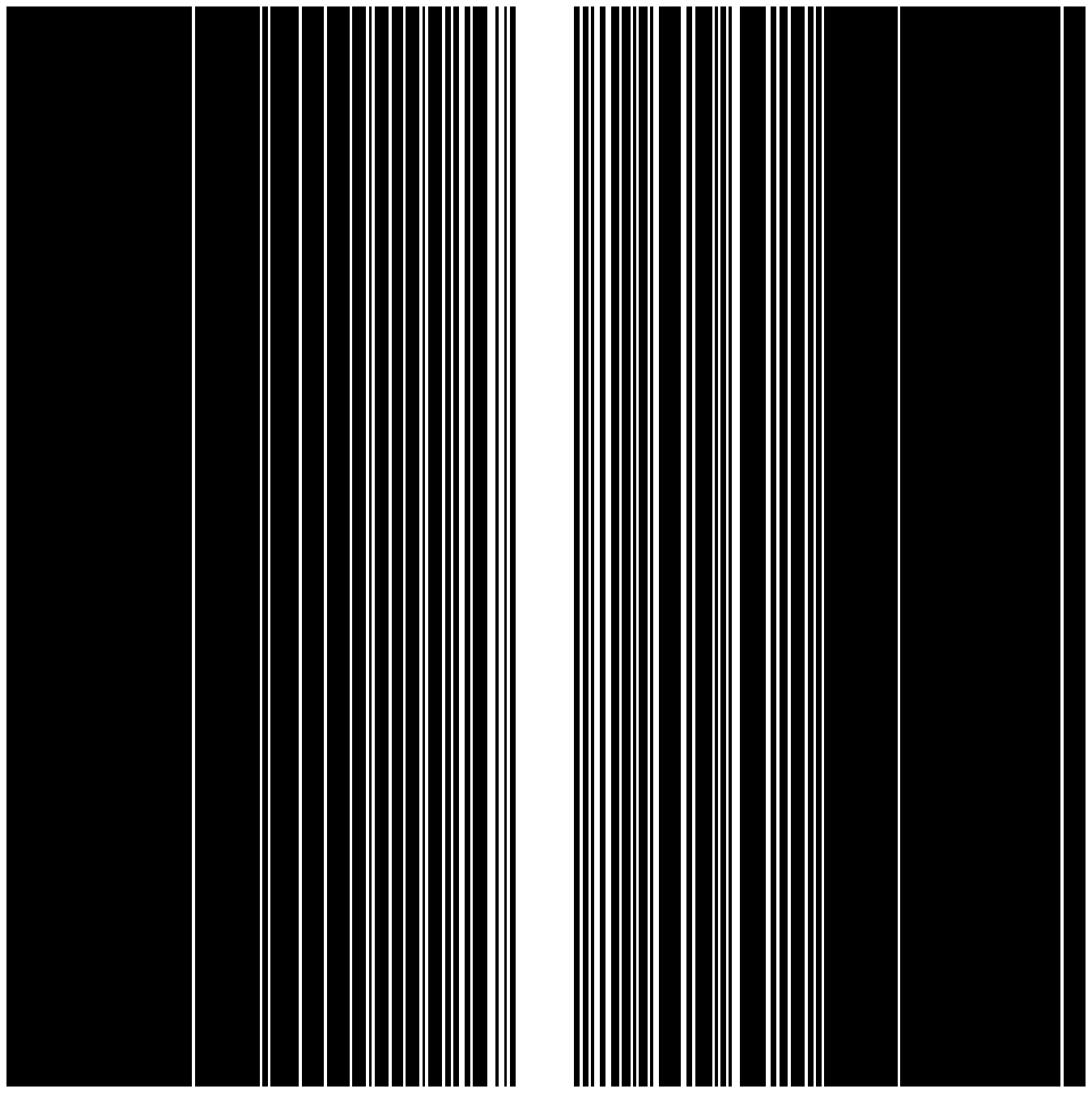}}
\caption{ Cartesian sampling mask for (left) $t$=1 and (right) subsequent frames. }
\label{fig:mask2}
\end{figure}

\begin{figure}
\centering
{\includegraphics[trim=10mm 0mm 5mm 5mm, clip=true, width=.5\textwidth]{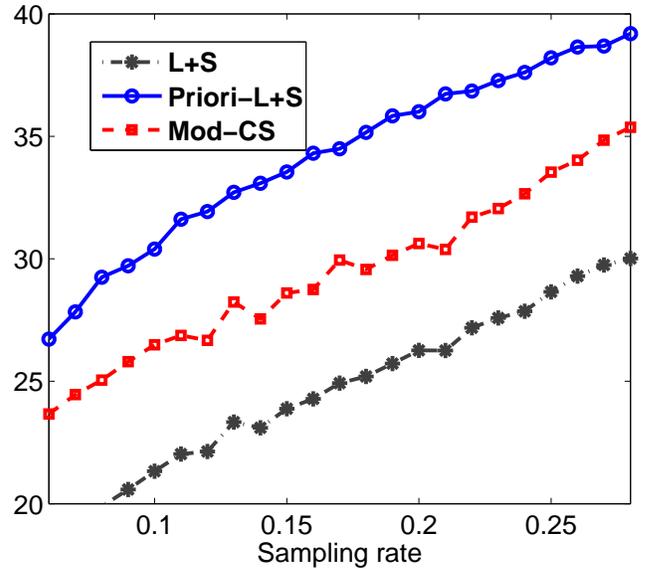}}
\caption{PSNR of the reconstructed images vs. sampling rate.}
\label{fig:res2D2}
\end{figure}

\section{EXPERIMENTAL RESULTS}
\label{XP}

To evaluate the performance of the proposed Priori $L+S$ method, we applied it to the reconstruction of dynamic 3D Cardiac volumes of size $256\times256\times14\times20$. The results are then  compared with that of L+S method  \cite{Otazo2013}, and also with Modified-CS method  \cite{Vaswani2010} (Mod-CS in figures 3 \&4). In all the experiments, we used a variable density Cartesian sampling
mask which in practice is less time consuming than random sampling.  However, to take the energy distribution of MR images in k-space  into account, we used a variable-density sampling with denser sampling near the center.
 Figure \ref{fig:mask2} shows the sampling masks used in these experiments with two different.  It should be noted that for the very first time-frame, since no priori information is available $\%50$ of the k-space samples are taken and  the sampling rate reported in figure 3 is for the successive frames.  Moreover, the sparse domain is assumed to be the Wavelet domain and the reconstruction
quality is measured using the Peak signal-to-noise ratio (PSNR).

\begin{figure}[!h]
\centering
\subfigure [ L+S]
{\includegraphics[trim=25mm 20mm 25mm 10mm, clip,width=0.155\textwidth]{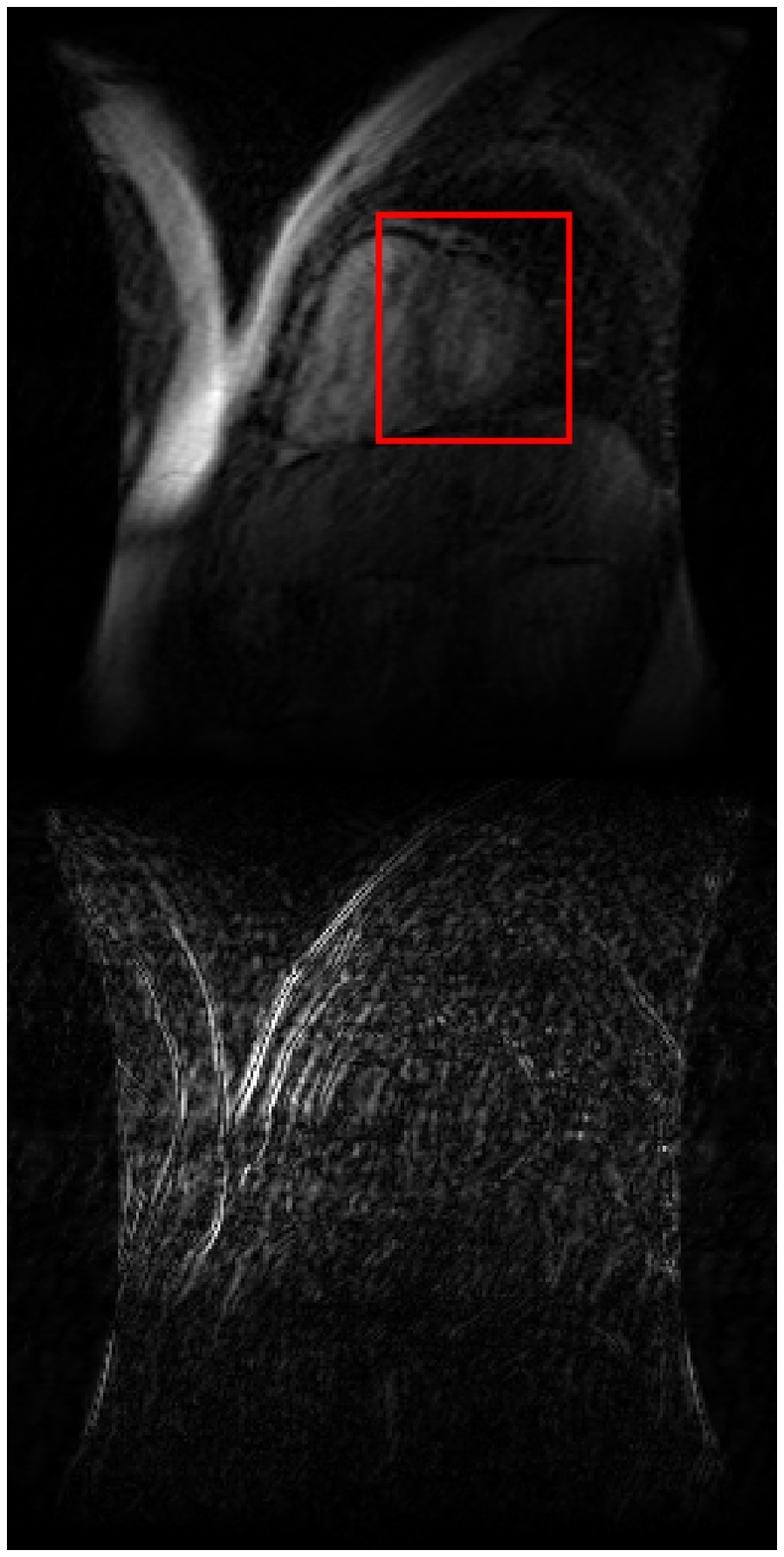}} 
\subfigure [Mod-CS]
{\includegraphics[trim=25mm 20mm 25mm 10mm, clip,width=0.155\textwidth]{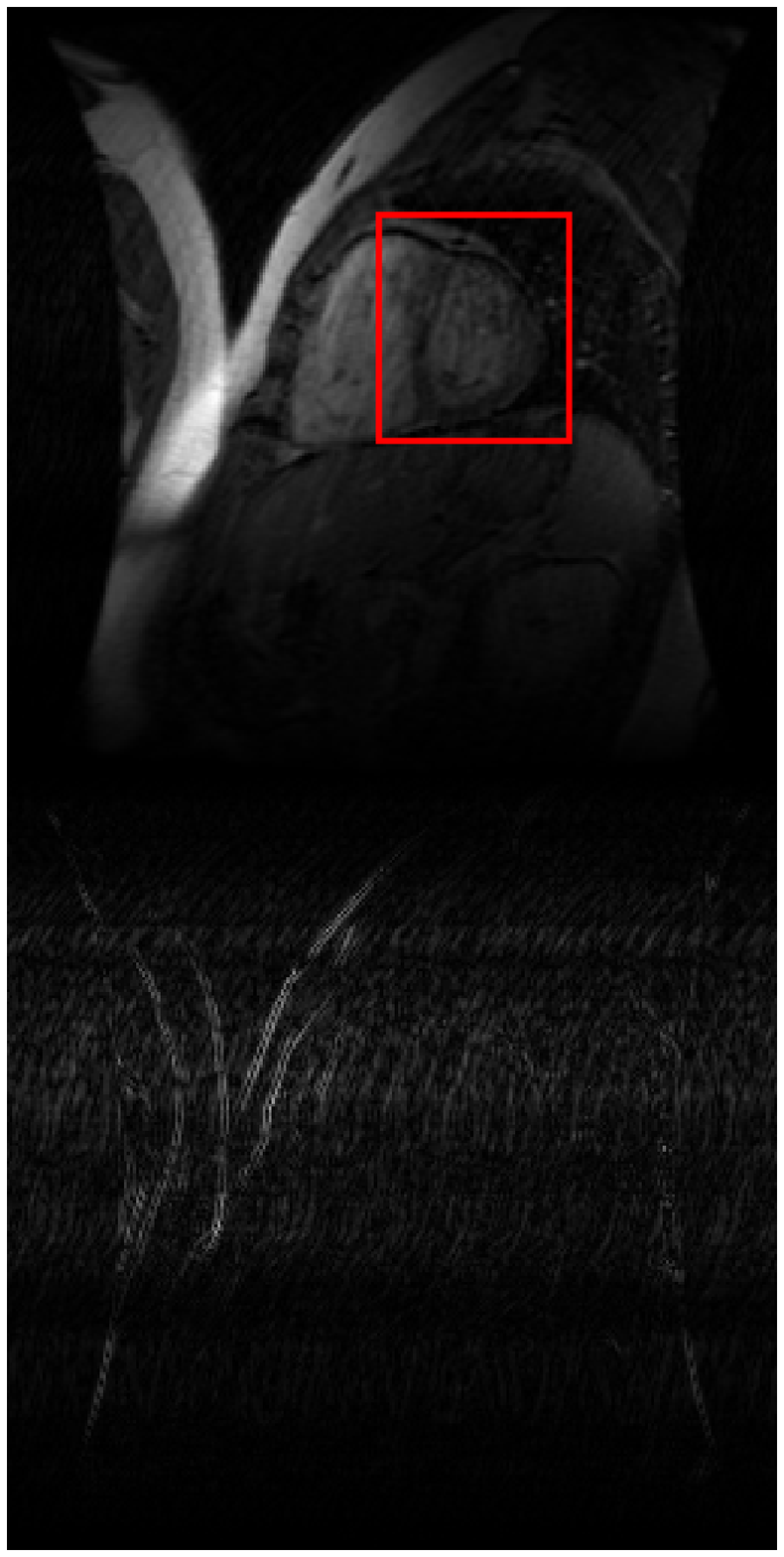}} 
\subfigure [Priori-L+S ]
{\includegraphics[trim=25mm 20mm 25mm 10mm, clip,width=0.155\textwidth]{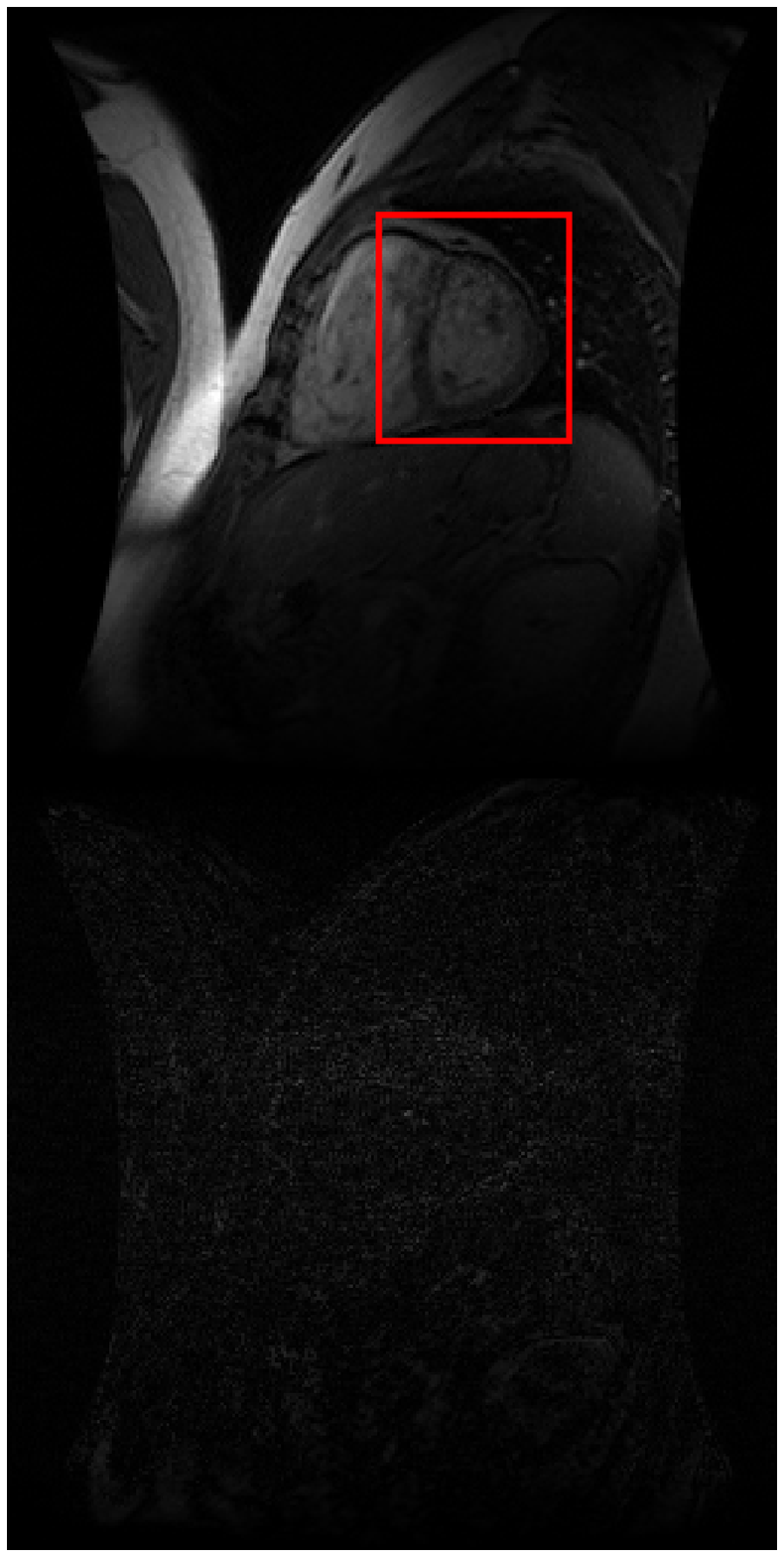}}\\
\caption{Comparison of the reconstructed images (1/7 of samples taken), together with the difference images that are amplified by a factor of 4. }
\label{fig:res-card}
\end{figure}

 Figure \ref{fig:res2D2} compares the average PSNR of the reconstructed volumes vs. percentage of the samples taken in the $k$-space.   It can be seen that our method consistently out-performs the others in terms of the improved PSNR. To compare the visual quality of the reconstructed images, figure \ref{fig:res-card} shows
a slice of the reconstructed volume using different methods together with the difference
images (reconstruction error) amplified by a factor of 4. It is evident that the reconstructed image using the Priori-L+S method is perceptually
better with less loss of details and significantly reduced reconstruction error.

\section{Conclusions}
\label{con}

In this paper, we presented a method which utilizes  a-priori knowledge for   high resolution and fast
reconstruction of dynamic 3D MRI image sequences from
undersampled k-space data. First, the problem of recovering a MRI images  as a sum of low-rank and sparse components ( $\L+ \S$) has been reformulated,   to incorporate the priori knowledge extracted from previous reconstructions. 
Then we proposed an iterative soft thresholding-based algorithm to efficiently
solve this minimization problem. To evaluate its performance, we used it to reconstruct
a time sequence of 3D cardiac MRI volumes from highly
undersampled  $k$-space data. Our experiments show that our 
proposed method is superior to the other state-of-the-art CS-based methods, in terms of both visual quality and improved
PSNR. Further investigation is still needed to study
the effect of the sparsifying transforms and sampling patterns
on the performance of the proposed Priori-L+S.

\bibliographystyle{IEEEbib}
\bibliography{refs}
\end{document}